\pdfoutput=1

\documentclass[11pt]{article}

\usepackage[preprint]{acl}

\usepackage{times}
\usepackage{latexsym}

\usepackage[T1]{fontenc}

\usepackage[utf8]{inputenc}

\usepackage{microtype}

\usepackage{inconsolata}

\usepackage{graphicx}
\usepackage{multirow}

\usepackage{array}
\usepackage{caption}
\usepackage{float}
\usepackage{stfloats}
\restylefloat{table}

\usepackage{placeins}

\usepackage{hyperref}

\usepackage[affil-it]{authblk}

\title{AI-based approach to burnout identification from textual data}

\author[1]{Marina Zavertiaeva}
\author[1]{Petr Parshakov}
\author[1]{Mikhail Usanin}
\author[1]{Aleksei Smirnov}
\author[1]{Sofia Paklina}
\author[1]{Anastasiia Kibardina}

\affil[1]{International Laboratory of Intangible-Driven Economy, \protect\\ \textit{National Research University Higher School of Economics, Russia}}

\begin{document}
\maketitle

\begin{abstract}
This study introduces an AI-based methodology that utilizes natural language processing (NLP) to detect burnout from textual data. The approach relies on a RuBERT model originally trained for sentiment analysis and subsequently fine-tuned for burnout detection using two data sources: synthetic sentences generated with ChatGPT and user comments collected from Russian YouTube videos about burnout. The resulting model assigns a burnout probability to input texts and can be applied to process large volumes of written communication for monitoring burnout-related language signals in high-stress work environments.

\end{abstract}

\section{Introduction}

Burnout syndrome, characterized by emotional, physical, and mental exhaustion due to chronic workplace stress that is not effectively managed (International Classification of Diseases, 11th Revision), is increasingly posing a serious problem for the economy. Burnout is accompanied by diminished energy, disengagement from work tasks, and decreased employee productivity. These issues have various adverse economic consequences, such as lower labor productivity, diminished performance outcomes, increased turnover costs, and a shrinking workforce.

Existing methods for identifying burnout primarily rely on self-reported surveys. The estimates derived from these surveys are prone to various issues, as they can be influenced by the respondent’s current mood and behavioral biases. Moreover, burnout is often identified only in its later stages, when individuals have already experienced a significant drop in mood and productivity. This delay complicates the timely detection of burnout at the individual level, making it harder to prevent and assess its economic impact. Artificial intelligence (AI), specifically natural language processing (NLP) techniques, can address these challenges in burnout detection by providing more objective and rapid evaluations with less influence from other psychological states. Additionally, AI can analyze large samples of individuals, enabling both individuals and organizations to take timely action. Nonetheless, the development of NLP-based methods for measuring burnout is still in its early stages, and existing approaches remain limited, particularly with respect to publicly described applications on non-English user-generated content. Therefore, the aim of this study is to develop a tool utilizing natural language processing (NLP) to detect burnout from textual data and to demonstrate its application on user-generated content, using a combination of real and synthetic texts.

To achieve this goal, the paper develops a methodology for identifying burnout using artificial intelligence methods, specifically NLP techniques. We fine-tune the RuBERT model for sentiment analysis using 22,994 sentences obtained from two types of data: synthetic data generated by ChatGPT and real user comments on YouTube videos about burnout.

This paper aims to contribute to the literature with its interdisciplinary approach, integrating modern advances in artificial intelligence methods to assess the psycho-emotional state of individuals using Russian-language data. First, we propose a method for detecting burnout using NLP techniques on Russian-language texts, specifically by fine-tuning the RuBERT model for sentiment analysis. Second, we provide a transparent labeling protocol grounded in ICD-11 and commonly used burnout inventories to support reproducibility and future extensions.

The paper begins with a brief discussion of the literature on the phenomenon of burnout and its measurement using AI-based methods. It then provides a description of the methodological part, which is devoted to the detection of burnout using NLP techniques based on Russian-language texts. The paper concludes with a discussion of the findings and limitations.

\section{Literature Review}

Burnout affects a significant number of workers across various industries and geographies. In 2019, the World Health Organization (WHO) included burnout in the 11th version of the International Classification of Diseases (ICD-11). According to ICD-11, burnout is “a syndrome conceptualized as resulting from chronic workplace stress that has not been successfully managed” (ICD-11, 2019). Maslach and Jackson’s three-factor model defines burnout through three components: emotional exhaustion, depersonalization, and reduced personal achievement (Maslach et al., 1997). Emotional exhaustion is characterized by feelings of overwhelm and emptiness; depersonalization is marked by a loss of motivation and alienation from clients or colleagues; and reduced personal achievement is manifested by diminished self-esteem and feelings of inadequacy regarding one’s work.

Organizations should be concerned about employee burnout because it can adversely affect business processes and organizational performance. First, burnout leads to decreased productivity (Dewa et al., 2014; Nayeri et al., 2009; Leitão et al., 2021). Second, burnout is associated with increased turnover, as employees experiencing burnout tend to seek new professional environments that offer more suitable working conditions and greater support for their psychological well-being (Du Plooy et al., 2010; Willard-Grace et al., 2019; Chen et al., 2019; Ivanovic et al., 2020). This can result in additional costs for companies related to hiring and training new employees, ultimately affecting organizational performance. Third, high levels of stress and psychological problems associated with burnout increase employee healthcare costs. These costs encompass both direct medical expenses and indirect costs, such as lost workdays due to illness and injury (De Beer et al., 2013; Melamed et al., 2006a; Melamed et al., 2006b; Shirom et al., 2005).

Empirical studies have primarily relied on survey data as the classical method for assessing employee burnout. Standardized questionnaires, such as the Maslach Burnout Inventory (MBI) (Maslach, 1986) and its extended version (Barnett et al., 1999), the Burnout Measure (BM) (Pines, 2005), and the Oldenburg Burnout Inventory (OLBI) (Demerouti et al., 2003), are widely utilized. In addition to questionnaires, researchers have attempted to identify various biomarkers of burnout, including salivary and blood cortisol levels, blood pressure, heart rate, cholesterol levels, C-reactive protein, and prolactin (Danhof-Pont, 2011; Bayes et al., 2021).

Currently, employing AI models to analyze employees’ textual communication represents a promising direction for measuring burnout. This approach traces its roots to the content analysis methods of the mid-20th century, pioneered by researchers such as Murray, McClelland, and Gottschalk. At that time, texts generated through free association or projective stimuli (e.g., Thematic Apperception Test) were used to assess psychological traits and states (Murray, 1943). Researchers categorized these texts according to relevant criteria and calculated frequencies using specialized coefficients (Schiff, 2017). Language – both spoken and written – serves not only as a diagnostic tool in clinical and personality psychology (Morales et al., 2017) but is also employed in training AI to create predictive tools in organizational psychology and managerial economics. This analysis utilizes texts from social media (Merhbene et al., 2022), interviews (Nath \& Kurpicz-Briki, 2021), and workplace communication (Mäntylä et al., 2016). Some studies investigate psycholinguistic markers, such as atypical use of emotive vocabulary (Kahn et al., 2007; Belz et al., 2022). However, research on burnout prediction using text analysis remains limited.

\section{Data \& Model Training}

To identify burnout based on Russian-language texts, the RuBERT model for sentiment analysis is utilized. We selected this model due to the perception that sentiment analysis is closely aligned with the burnout identification problem. RuBERT is a Bidirectional Encoder Representations from Transformers (BERT) model that has been trained on the RuSentiment dataset (Rogers et al., 2018) for the sentiment analysis of Russian-language social media publications. The model is fine-tuned using two types of data:

\begin{enumerate}
    \item \textit{Synthetic data generated using ChatGPT.} ChatGPT 3.5 Turbo was utilized to generate a sample containing sentences that both exhibit and do not exhibit signs of burnout. The prompt requested 10 sentences indicating signs of burnout, along with 10 sentences that do not show signs of burnout, based on the characteristics of the hypothetical speaker. To enhance diversity in the output from GPT, several personal characteristics were varied in the prompt: gender (male/female), age (young/middle-aged/old), job experience (with/without), job position (executive/subordinate), communication method (verbal/written), communication type (professional/casual), and professional sphere. This resulted in a total of 3,264 unique prompts. Following data preparation, a total of 32,916 sentences indicating burnout and 32,171 sentences without burnout were obtained. On average, the sentences in both samples were 97 characters long.
    \item \textit{User comments on videos about burnout on YouTube.} Comments were collected from five popular Russian YouTube videos\footnote{
Emotional Burnout: How to Recognize and How to Cope | My 10 Tips. Alexandra Orlova. \href{https://www.youtube.com/watch?v=O1K6Cr-qkpU}{https://www.youtube.com/watch?v=O1K6Cr-qkpU}

Sadness, melancholy, burnout: how to recognize and overcome depression? / Editorial \href{https://www.youtube.com/watch?v=qh8wghUzMOM&list=PLYUIc_t4fCOeE8cYK2LkA3YbrtLWxABB8&index=2&pp=iAQB}{https://www.youtube.com/watch?v=qh8wghUzMOM}

Emotional burnout. Treatment by stages. Sabbatical. \href{https://www.youtube.com/watch?v=IHZEqujLtFU&list=PLYUIc_t4fCOeE8cYK2LkA3YbrtLWxABB8&index=3&pp=iAQB}{https://www.youtube.com/watch?v=IHZEqujLtFU}

BURNOUT. For those who have LOST motivation and strength. M Generation. \href{https://www.youtube.com/watch?v=mW4eYX2XFxg&list=PLYUIc_t4fCOeE8cYK2LkA3YbrtLWxABB8&index=4&pp=iAQB}{https://www.youtube.com/watch?v=mW4eYX2XFxg}

BURNOUT. How to find balance and not ruin your life because of work. \href{https://www.youtube.com/watch?v=zoMHxsbEuX8}{https://www.youtube.com/watch?v=zoMHxsbEuX8}
} about burnout. The comments were preprocessed, which involved removing meaningless and short comments, and then tokenized. We utilized Chat-GPT 3.5 Turbo to classify sentences for indications of burnout. To address ambiguity in certain sentences, we also fine-tuned RuBERT using synthetic data specifically for burnout detection. The final dataset from this source comprises 7,888 sentences indicating burnout and 8,505 sentences without burnout. On average, the comments are 82 characters long.
\end{enumerate}

Out of 16,393 YouTube comments from the cleaned dataset, 1,601 comments exhibited discrepant assessment results (i.e., the pre-assessment by GPT indicated that “the comment is likely to indicate burnout,” while the BERT assessment after the first training iteration classified it as “unlikely to indicate burnout,” and vice versa). Such ambiguous comments were labeled manually based on the parameters and criteria detailed in Table 1.

\FloatBarrier
\begin{table}[h]
\centering
\resizebox{\textwidth}{!}{  % Resize the table to fit within the page width
\begin{tabular}{lp{2cm}p{7cm}p{8cm}}
\hline
\textbf{Parameters} & \textbf{Values} & \textbf{Criteria} & \textbf{Example comments} \\
\hline
\multirow{7}{*}{\textbf{Burnout indicators}} & \multirow{4}{*}{Present} & Comment provides self-report on burnout experience, commenter uses the word “burnout” or its derivatives explicitly. & "I'm \textit{burning out} for the second time, and it feels just as sudden as the first time. Out of many videos about burnout, yours is the coolest and most informative." \\
 &  & Comment contains keywords matching the keywords indicating burnout in the 11th revision of  International Classification of Diseases (ICD-11), Maslach Burnout Inventory (MBI), Boiko Burnout Inventory (BBI). & “A new startup, a year of quarantine, and a \textit{series of }big problems and \textit{stresses} have taken their toll.”

"At the same time, the work itself might not have been all that difficult, but the importance I attached to it and the degree of \textit{dissatisfaction I felt with myself} completely consumed me." \\
 &  & Comment contains words synonymous to the keywords indicating burnout in ICD-11, MBI, BBI. & “I worked in a call center for 3 months, and it helped to imagine that it wasn't people calling, but \textit{some schizo-robots}.” \\
 &  & Comment implicates burnout experience by conveying the meaning similar to the meaning of utterances in MBI/ BBI and matching the ICD-11 burnout context; there are no explicit matches with ICD-11, MBI, BBI though. & “I recently asked myself — why do I go to work?”

"Once I didn't want to go to work so badly that I woke up in the morning with a fever."

"How to relax when the work schedule is 6/1, with no holiday breaks?" \\
\cline{3-4}
 & \multirow{2}{*}{Not present} & Comment provides self-report on having no burnout experience, commenter uses the word “burnout” or its derivatives explicitly. & "If you really love your work, you will never experience \textit{burnout} (I never did)." \\
 &  & Comment implicates no burnout experience by conveying the meaning of bewilderment, devaluation or denial in response to comments with present burnout indicators. & "Is it even possible for a person who is talented and passionate about their work to experience burnout?"

"If you experience unpleasant sensations, don't focus on them; redirect your attention."

"With all due respect: tell about burnout to a person rotting from gangrene, or to someone blinded by a tumor that has taken up a third of their brain, or to someone burned over 20-30 percent of their total body surface."

"Burnout is a 'product' that you 'bought'." \\
\cline{3-4}
 & N/A & Lack of information for adequate assessment. & "The speaker laughs endlessly while discussing serious matters — couldn't watch the video out to the end because of that.” \\
\hline
\multirow{3}{*}{\textbf{Time relevance indicators}} & Present & Commenter refers to acute condition. & "Right now, the topic is very pertinent to me; I recognize myself 100\% in your description of burnout.” \\
\cline{3-4}
 & Not present & Commenter refers to past condition. & "I was experiencing the same symptoms for the whole year, back in 2008." \\
\cline{3-4}
 & N/A & Lack of information for adequate assessment / Inapplicable. & "It is harder to recover than to take preventive measures!" \\
\hline
\multirow{4}{*}{\textbf{Relevance}} & Relevant & Commenter refers to her own work situation. & "I am an interior designer and I give more emotions and joy than I receive." \\
\cline{3-4}
 & \multirow{3}{*}{Irrelevant} & Commenter refers to someone else’s work situation (including the situations of her relatives, friends, colleagues or of an author, who created the video). & "Apparently, my dad has at least stage three after so many years of working as a dentist without a vacation. He used to be very friendly and loved his patients, and they loved him. Now he can sit and play on his phone while the patient is waiting." \\
 &  & Commenter refers to a different condition (e.g. depression or PTSD). & “According to Beck's Depression Inventory, it turned out that I have severe depression.” \\
 &  & Commenter shares off-topic opinions. & "Evgenia, your posture is very poor; please take care of your back." \\
\hline
\multirow{2}{*}{\textbf{Confidence}} & 1 & Conclusions resulting from the manual assessment seem adequate. & “Every morning on my way to work I just want to give up everything (everyone is so annoying!), go nowhere and live alone in the woods. I sob a little and still get to work.” \\
\cline{3-4}
 & 0 & Assessment is impeded, and the resulting conclusions might not be adequate. & "I want a break - You haven't worked enough!" \\
\hline
\end{tabular}
}
\caption{Parameters and criteria of manual labeling }
\end{table}
\FloatBarrier

The identification of burnout was guided by the interpretation of burnout proposed in ICD-11 (World Health Organization, 2022). Additionally, we relied on burnout assessment tools commonly used among Russian-speaking populations: the Maslach Burnout Inventory (MBI) (Maslach et al., 1997) and the Boiko Burnout Inventory (BBI) (Boiko, 2013). All comments that were content-wise aligned with the construct of burnout as reflected in ICD-11, MBI, or BBI (Confidence = 1) were labeled as “Likely to indicate burnout.” Comments describing past burnout experiences were excluded, as the intention was to ultimately train the model to screen for acute conditions. Relevant comments lacking burnout indicators, along with irrelevant comments (Confidence = 1), were labeled as “Unlikely to indicate burnout.” For the subsequent model training, only those manually labeled comments that exhibited the specific combination of parameter values described in Table 2 were used.

\FloatBarrier
\begin{table*}[bp]
\centering
\begin{tabular}{p{2cm}p{1.5cm}p{1.5cm}p{1.5cm}p{1.5cm}p{4cm}}
\hline
\textbf{Parameter →}

\textbf{Labeling ↓} & \textbf{Burnout indicators} & \textbf{Time relevance indicators} & \textbf{Relevance} & \textbf{Confidence} & \textbf{← Parameter}

\textbf{Example ↓} \\
\hline
\textit{“The comment is likely to indicate burnout”.} & Present & Present & Relevant & 1 & “Because of the stress at work, I’ve come down with a chronic disease. I wish someone would just come and drag me out of work. At times I even have thoughts like ‘wouldn’t it be nice to wind up in hospital?’ because it’s somewhat unreal to cope with on your own, even though you recognise the problem.” \\
\hline
\textit{“The comment is unlikely to indicate burnout”.} & Not present & N/A & Irrelevant & 1 & “There’s no need to make yourself crazy for nothing, seriously. Everyone concentrates so much on negative moments, and there’s really a lot of those, but here's another truth — try and learn to find joy in every moment, be grateful for every moment.” \\
\hline
\end{tabular}
\caption{Configuration of parameter values of the manually labeled comments used for the subsequent model training. }
\end{table*}
\FloatBarrier

Examples of generated sentences and annotated comments are presented in Table 3.

\begin{table*}[h]
\centering
\begin{tabular}{>{\raggedright\arraybackslash}p{0.2\linewidth}>{\raggedright\arraybackslash}p{0.35\linewidth}>{\raggedright\arraybackslash}p{0.35\linewidth}}
\hline
\textbf{Source of the data} & \textbf{Burnout} & \textbf{Neutral} \\
\hline
Synthetic & “Dear colleagues, I've been feeling pretty exhausted lately. Could you please share your tips for managing stress?”

“I apologize for the delay in responding. The last few days have been very stressful due to the large amount of work.” & “My professional confidence and experience help me handle difficult work situations with ease and professionalism.”

“Hey, guys! How are you guys doing? I'm energized and ready to get to work today.” \\
YouTube comments & “My empathy started to burn out, I realized I'd become indifferent to people”

“Since June, I've been kind of lethargic, sad, unproductive, and annoyed all the time”  & “Your words were a push to change and change a lot!”

“Thanks to you, I started learning French, began considering education abroad, and found inspiration.” \\
\hline
\end{tabular}
\caption{Examples of the sentences used for training the model}
\end{table*}

To maintain a balanced composition between synthetic and real data, we included only 5,000 randomly sampled generated sentences, making up approximately 20\% of the final dataset. This approach ensured a predominance of real data while still integrating a representative subset of synthetic sentences.

All layers of the RuBERT model were frozen except for the last fully connected layer (classifier), and the model was fine-tuned using a combination of synthetic data and YouTube comments. The characteristics of the final training dataset are presented in Table 4.

\begin{table*}[h]
\centering
\renewcommand{\arraystretch}{1.5}

\begin{tabular}{>{\raggedright\arraybackslash}p{0.1\linewidth}>{\raggedright\arraybackslash}p{0.2\linewidth}>{\raggedright\arraybackslash}p{0.2\linewidth}>{\raggedright\arraybackslash}p{0.2\linewidth}>{\raggedright\arraybackslash}p{0.2\linewidth}}
\hline
  & \textbf{Type of data} & \textbf{Number of sentences} & \textbf{Average number of characters in a sentence} & \textbf{Average number of words in a sentence} \\
\hline
\textbf{Burnout} & Synthetic & 1,981 & 93.73 & 14.53 \\
 & YouTube comments with GPT labelling & 6,327 & 91.16 & 15.06 \\
 & YouTube comments with manual labelling & 566 & 81.89 & 13.68 \\
\textbf{Neutral} & Synthetic & 2,015 & 101.66 & 15.36 \\
 & YouTube comments with GPT labelling & 6,772 & 73.46 & 11.62 \\
 & YouTube comments with manual labelling & 734 & 77.28 & 12.51 \\
\hline
\textbf{Total} &  & 18,395 & 79.72 & 12.48 \\
\hline
\end{tabular}
\caption{Characteristics of the training dataset}
\end{table*}

The model was trained for five epochs with a linearly decreasing learning rate, starting from 5*10\textsuperscript{-5}. The batch size remained fixed at 256 throughout the training process. The training visualization is presented in Figure 1.

\begin{figure*}[h]
    \centering
    \includegraphics[width=1\linewidth]{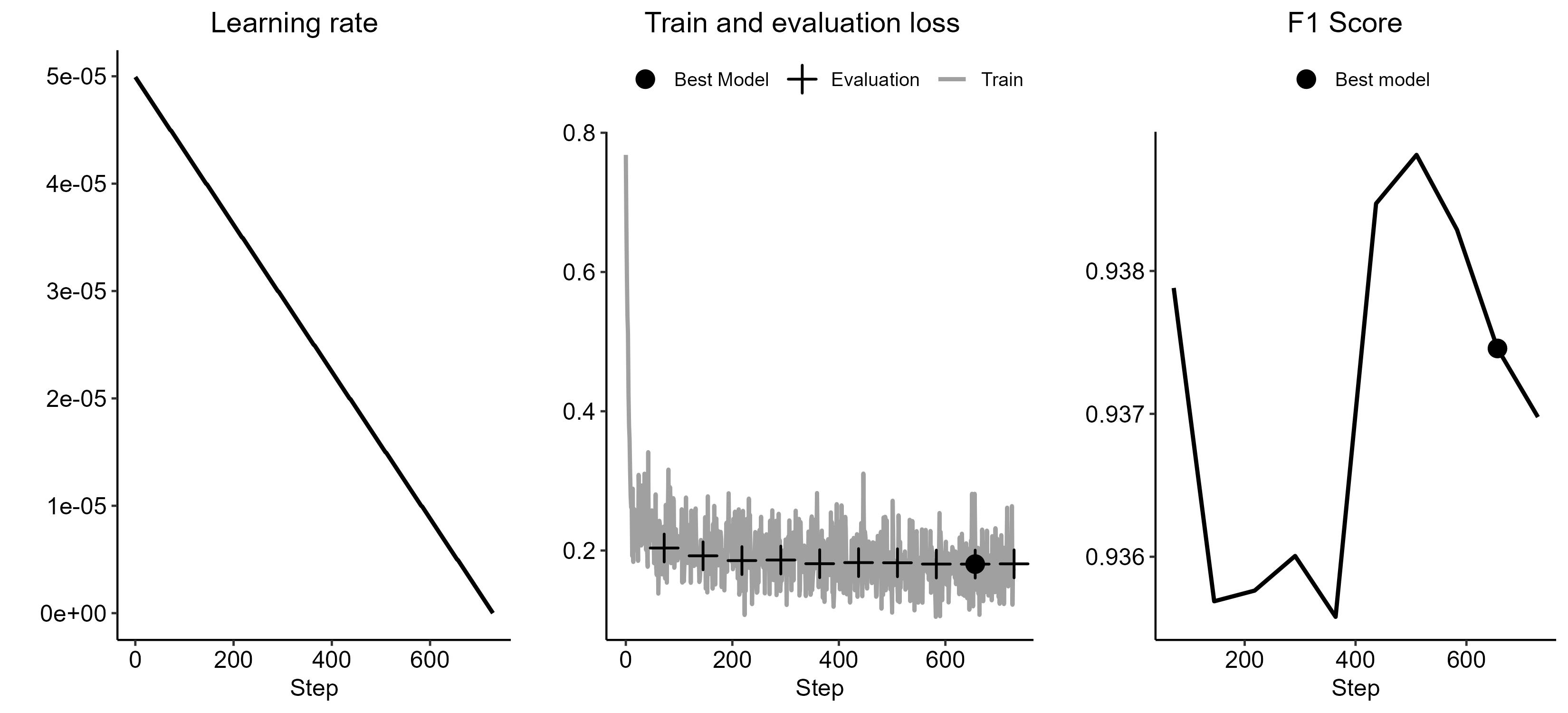}
    \caption{Training and evaluation metrics}
    \label{fig:enter-label}
\end{figure*}

At the end of the fifth epoch, the model achieved an accuracy of 0.940, with precision = 0.931, recall = 0.944, F-score = 0.937, and AUC-ROC = 0.980, demonstrating robust performance across multiple evaluation criteria. The ROC curve is displayed in Figure 2. Evaluation metrics and the confusion matrix are detailed in the Appendix.

\begin{figure}[h]
    \centering
    \includegraphics[width=1\linewidth]{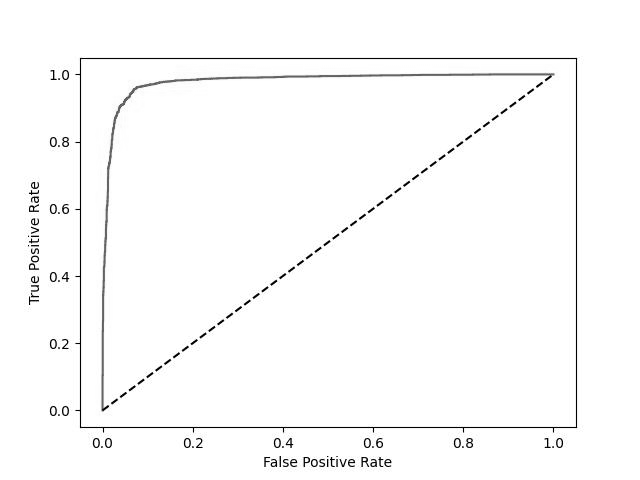}
    \caption{ROC Curve}
    \label{fig:enter-label-roc}
\end{figure}

It can be concluded that the model demonstrates strong performance across various evaluation metrics. However, further validation on external datasets and consideration of potential biases are necessary to ensure the robustness and generalizability of the model’s performance in real-world scenarios.

\section{Discussion}

This study introduces an approach utilizing artificial intelligence, specifically natural language processing (NLP), to detect and analyze burnout through textual data. The BERT language model was fine-tuned to identify indications of burnout using a combination of synthetic texts and real-world user-generated content, with experiments demonstrated on Russian-language data.

The implications of this study are significant across multiple fields. Firstly, the developed methodology offers a more objective and scalable approach to detecting burnout compared to traditional self-report surveys, which are often influenced by biases and limitations. By harnessing NLP and AI, organizations can effectively monitor and swiftly identify signs of burnout, enabling timely interventions.

Secondly, this research contributes to the broader conversation on workplace mental health, advocating for increased awareness and proactive management of burnout to uphold organizational performance. As workplaces evolve amidst mounting pressures and challenges, the methodologies and insights from this study provide valuable tools for organizations across diverse industries to understand and mitigate the risks associated with burnout.

\section{Conclusion}

In conclusion, the AI-based approach explored in this paper presents practical opportunities for monitoring burnout in Russian-language textual communication. The proposed methodology can support early detection and inform organizational interventions aimed at improving well-being and sustaining effective performance in high-stress environments.

\section{Limitations}

Despite these findings, this study faces several limitations. Firstly, while the AI-based methodology demonstrated strong performance on the curated training and evaluation data, further validation is necessary on external datasets to ensure robustness and generalizability across domains, demographics, and communication styles. Secondly, labeling relies on a combination of automated (LLM-based) and manual procedures; future research should assess sensitivity to labeling noise and evaluate alternative annotation protocols. Addressing these limitations would strengthen the applicability of the proposed methodology for monitoring burnout and enhancing workplace mental health in real-world scenarios.

\section{References}

Barnett, R. C., Brennan, R. T., \& Gareis, K. C. (1999). A closer look at the measurement of burnout 1. \textit{Journal of Applied Biobehavioral Research}, \textit{4}(2), 65-78.

Belz, F. F., Adair, K. C., Proulx, J., Frankel, A. S., \& Sexton, J. B. (2022). The language of healthcare worker emotional exhaustion: A linguistic analysis of longitudinal survey. \textit{Frontiers in Psychiatry, 13}. \href{https://doi.org/10.3389/fpsyt.2022.1044378}{https://doi.org/10.3389/fpsyt.2022.1044378}

Boyko, V. V. (1996). \textit{Diagnostics of the level of emotional burnout}. \href{http://new.academy.edu.by/files/documents/VRVUO/Boiko-Ilin_Diagnostika.pdf}{http://new.academy.edu.by/files/documents/VRVU\\O/Boiko-Ilin\_Diagnostika.pdf} [in Rus.]

Boiko, V. V. (2013). Sindrom «emocionalnogo vygoraniya» v professionalnom obshchenii [Emotional burnout syndrome in professional communication]. \textit{Piter.} [in Rus.]

Chen, X., Ran, L., Zhang, Y., Yang, J., Yao, H., Zhu, S., \& Tan, X. (2019). Moderating role of job satisfaction on turnover intention and burnout among workers in primary care institutions: a cross-sectional study. \textit{BMC public health}, \textit{19}, 1-10.

De Beer, L., Pienaar, J., \& Rothmann Jr, S. (2013). Linking employee burnout to medical aid provider expenditure. \textit{South African Medical Journal}, \textit{103}(2), 89-93.

Demerouti, E., Bakker, A. B., Vardakou, I., \& Kantas, A. (2003). The convergent validity of two burnout instruments: A multitrait-multimethod analysis. \textit{European Journal of Psychological Assessment}, \textit{19}(1), 12.

Dewa, C.S., Loong, D., Bonato, S. \textit{et al.} How does burnout affect physician productivity? A systematic literature review. \textit{BMC Health Serv Res} \textbf{14}, 325 (2014). \href{https://doi.org/10.1186/1472-6963-14-325}{https://doi.org/10.1186/1472-6963-14-325}

Du Plooy, J., \& Roodt, G. (2010). Work engagement, burnout and related constructs as predictors of turnover intentions. \textit{SA journal of Industrial Psychology}, \textit{36}(1), 1-13.

\href{https://www.gazeta.ru/social/news/2023/03/20/20013577.shtml}{https://www.gazeta.ru/social/news/2023/03/20/2\\0013577.shtml.}

ICD-11 for Mortality and Morbidity Statistics. – 2023. URL: \href{https://icd.who.int/browse11/l-m/ru\#/http://id.who.int/icd/entity/129180281}{https://icd.who.int/browse11/l-m/ru\#/http://id.who.int/icd/entity/129180281}

Ivanovic, T., Ivancevic, S., \& Maricic, M. (2020). The relationship between recruiter burnout, work engagement and turnover intention: Evidence from Serbia. \textit{Engineering Economics}, \textit{31}(2), 197-210.

Kahn, J. H., Tobin, R. M., Massey, A. E., \& Anderson, J. A. (2007). Measuring emotional expression with the Linguistic Inquiry and Word Count. \textit{The American Journal of Psychology}, 120(2), 263–286.

Korsakov N. (2023, March 20). Sber told about AI-model of HR-platform "Pulse". \textit{Gazeta.ru}. (in Russian)

Leitão, J., Pereira, D., \& Gonçalves, Â. (2021). Quality of work life and contribution to productivity: Assessing the moderator effects of burnout syndrome. \textit{International Journal of Environmental Research and Public Health}, \textit{18}(5), 2425.

Malayan, K. R., Milokhov, V. V., Minko, V. M., Rusak, O. N., Faustov, S. A., Tsaplin, V. V., \& Tsvetkova, A. D. (2014). Scientific commentary to the legislation on special assessment of labor conditions. \textit{SPb.: IP Pavlushkina VN}. (in Russian)

Mäntylä, M., Adams, B., Destefanis, G., Graziotin, D., \& Ortu, M. (2016). Mining valence, arousal, and dominance: Possibilities for detecting burnout and productivity? Proceedings of the 13th International Conference on Mining Software Repositories, 247–258. \href{https://doi.org/10.1145/2901739.2901752}{https://doi.org/10.1145/2901739.2901752}

Maslach, C. (1986). Stress, burnout, and workaholism.

Maslach, C., Jackson, S. E., \& Leiter, M. P. (1997). Maslach Burnout Inventory: Third edition. In C. P. Zalaquett \& R. J. Wood (Eds.), Evaluating stress: A book of resources. \textit{Scarecrow Education}, pp. 191–218.

Melamed, S., Shirom, A., Toker, S., \& Shapira, I. (2006). Burnout and risk of type 2 diabetes: a prospective study of apparently healthy employed persons. \textit{Psychosomatic medicine}, \textit{68}(6), 863-869.

Melamed, S., Shirom, A., Toker, S., Berliner, S., \& Shapira, I. (2006). Burnout and risk of cardiovascular disease: evidence, possible causal paths, and promising research directions. \textit{Psychological bulletin}, \textit{132}(3), 327.

Merhbene, G., Nath, S., Puttick, A. R., \& Kurpicz-Briki, M. (2022). BurnoutEnsemble: Augmented Intelligence to Detect Indications for Burnout in Clinical Psychology. \textit{Frontiers in Big Data}, 5. \href{https://www.frontiersin.org/articles/10.3389/fdata.2022.863100}{https://www.frontiersin.org/articles/10.3389/fdata.\\2022.863100}

Morales, M., Scherer, S., \& Levitan, R. (2017). A Cross-modal Review of Indicators for Depression Detection Systems. B. K. Hollingshead, M. E. Ireland, \& K. Loveys, Proceedings of the Fourth Workshop on Computational Linguistics and Clinical Psychology—From Linguistic Signal to Clinical Reality (pp. 1–12). \textit{Association for Computational Linguistics}. \href{https://doi.org/10.18653/v1/W17-3101}{https://doi.org/10.18653/v1/W17-3101}

Murray, H. A. (1943). \textit{Thematic apperception test}. Harvard University Press.

Nath, S., \& Kurpicz-Briki, M. (2021). BurnoutWords—Detecting Burnout for a Clinical Setting. \textit{Machine Learning Techniques and Data Science}, 177–191. \href{https://doi.org/10.5121/csit.2021.111815}{https://doi.org/10.5121/csit.2021.111815}

Nayeri, N. D., Negarandeh, R., Vaismoradi, M., Ahmadi, F., \& Faghihzadeh, S. (2009). Burnout and productivity among Iranian nurses. \textit{Nursing \& health sciences}, \textit{11}(3), 263-270.

Pines, A.M. (2005), The burnout measure, short version. \textit{International Journal of Stress Management}, 12(1), 78-88.

Rogers, A., Romanov, A., Rumshisky, A., Volkova, S., Gronas, M., \& Gribov, A. 2018. \href{https://aclanthology.org/C18-1064}{RuSentiment: An Enriched Sentiment Analysis Dataset for Social Media in Russian}. In \textit{Proceedings of the 27th International Conference on Computational Linguistics}, pages 755–763, Santa Fe, New Mexico, USA. Association for Computational Linguistics.

Schiff, B. (2017). Turning to Narrative. In B. Schiff (Ed.), A New Narrative for Psychology (p. 43-70). \textit{Oxford University Press}. \href{https://doi.org/10.1093/oso/9780199332182.003.0004}{https://doi.org/10.1093/oso/9780199332182.003.0004}

Shirom, A., Melamed, S., Toker, S., Berliner, S., \& Shapira, I. (2005). Burnout and health review: Current knowledge and future research directions. \textit{International review of industrial and organizational psychology}, \textit{20}(1), 269-308.

Willard-Grace, R., Knox, M., Huang, B., Hammer, H., Kivlahan, C., \& Grumbach, K. (2019). Burnout and health care workforce turnover. \textit{The Annals of Family Medicine}, \textit{17}(1), 36-41.

World Health Organization. (2022). ICD-11: International classification of diseases (11th revision). \href{https://icd.who.int/ru}{https://icd.who.int/ru}

Yanna, L., Yang, S., \& Lin, Z. Analysis of the Impact of Job Burnout on Quality and Economic Benefits of Enterprises. \textit{Journal of Economics, Management and Trade}, 29(9), 23-38.

Yaroshenko, E. I. (2019). Application of eye-tracking technology to identify socio-psychological features of emotional burnout of personality. \textit{Organizational Psychology}, 9(1), Article 1. (in Russian)

\section{Appendix}
\label{sec:appendix}

\centering

\begin{tabular}{l l}
\hline
Metric & Value \\
\hline
Accuracy & 0.940 \\
Precision & 0.931 \\
Recall & 0.944 \\
F-score & 0.937 \\
AUC-ROC & 0.980 \\
\hline
\end{tabular}

\captionof{table}{Evaluation metrics}

\bigskip

\begin{tabular}{l l l}
\hline
  & Predicted Negative & Predicted Positive \\
\hline
Actual Negative & TN = 2,246 & FP = 154 \\
Actual Positive & FN = 125 & TP = 2074 \\
\hline
\end{tabular}

\captionof{table}{Confusion matrix}

\end{document}